\documentclass{article}
\usepackage[utf8]{inputenc}
\usepackage[margin=2cm]{geometry}
\usepackage{fullpage,enumitem,amssymb,amsmath,amsthm,xcolor,cancel,gensymb,hyperref,graphicx}
\usepackage{bbm}

\usepackage{indentfirst}

\usepackage[ruled,vlined]{algorithm2e}

\usepackage{biblatex}
\newtheorem{thm}{Theorem}
\newtheorem{lem}[thm]{Lemma}

\newtheorem{cor}[thm]{Corollary}
\newtheorem{prop}{Proposition}

\DeclareMathOperator*{\argmax}{arg\,max}

\DeclareMathOperator*{\Var}{\text{Var}}
\DeclareMathOperator*{\Cov}{\text{Cov}}
\DeclareMathOperator*{\bbE}{\mathbb{E}}
\DeclareMathOperator*{\bbR}{\mathbb{R}}

\DeclareMathOperator*{\bbP}{\mathbb{P}}

\DeclareMathOperator*{\calA}{\mathcal{A}}

\DeclareMathOperator*{\calF}{\mathcal{F}}

\DeclareMathOperator*{\calH}{\mathcal{H}}
\DeclareMathOperator*{\calL}{\mathcal{L}}
\DeclareMathOperator*{\calM}{\mathcal{M}}

\DeclareMathOperator*{\calE}{\mathcal{E}}

\title{RTB Formulation Using Point Process}

\author{
    Seong Jin Lee \\
    \texttt{slee7@unc.edu} \\
    University of North Carolina, Chapel Hill
    \and
    Bumsik Kim \\
    \texttt{bumsik@moloco.com} \\
    Moloco Inc.
}

\begin{document}

\maketitle

\begin{abstract}
    We propose a general stochastic framework for modelling repeated auctions in the Real Time Bidding (RTB) ecosystem using point processes. The flexibility of the framework allows a variety of auction scenarios including configuration of information provided to player, determination of auction winner and quantification of utility gained from each auctions. We propose theoretical results on how this formulation of process can be approximated to a Poisson point process, which enables the analyzer to take advantage of well-established properties. Under this framework, we specify the player's optimal strategy under various scenarios. We also emphasize that it is critical to consider the joint distribution of utility and market condition instead of estimating the marginal distributions independently. \\
\end{abstract}

\section{Introduction}

With the rapid growth of the digital advertisement industry, programmatic advertisement became a crucial part of the industry. A key component of the programmatic display advertisement is the Real Time Bidding (RTB) where the supply-side platform (SSP) puts an ad-inventory on auction and the demand-side platforms (DSP) computes the potential value of the inventory and submits a bid according to the estimated value from the buyer-perspective to win the advertising opportunity.

Many studies have been conducted to propose the optimal strategies for each of the participant in this ecosystem. Some approaches uses classical auction theories from game-theoretical views\cite{aggarwal2019autobidding}, \cite{balseiro2021robust}, \cite{nedelec2022learning}, \cite{nedelec2019learning} which considers the strategies of SSP and the game between DSP's and the SSP. In this paper, we focus on the perspective of the DSP, where the player participates as a buyer in the auction.

The first step of addressing this optimization problem is to defining the objective. Common objectives of display advertising includes impression, click, conversion and revenue from the user driven by the ad. Traditional viewpoint of this problem is in the profit maximization problem, where valuations of each objectives are predefined. Because it is not straight forward to define the valuation of diverse objectives, another objective is to maximize the objective with respect to a budget constraint. Some literature formulate the strategy as an linear programming of integer programming \cite{chen2011real}, \cite{aggarwal2019autobidding} which considers fixed number of auctions and solves how to allocate the budget to maximize value. Some recognizes the problem in an optimal control point of view \cite{fernandez2017optimal}, \cite{karlsson2020feedback}, \cite{jiang2014bidding}, \cite{nuara2022online} to optimally allocate the budget over time. Another point of view uses Markov Decision Process point of view \cite{gummadi2013optimal} to use Reinforcement Learning theory. In this paper, we aim to provide a general framework that models abstract objectives, which gives flexibility to adapt diverse objectives that are commonly used in the industry.

In either formulation of the optimization problem, there are two key components auction participant needs to take account of. The first component is measuring the quantity of the ad opportunities. As the source of the ad inventories are from different users from worldwide acting in a random fashion, its occurrence is naturally stochastic. For effective design of spending control in this stochastic environment, it is crucial to understand its behaviour, and to model it correctly. There has been attempts to understand the ecosystem as a homogeneous marked Poisson process \cite{fernandez2017optimal}, but in this paper we model the ecosystem as a general point process. In section 2, we aim to model this natural landscape of bid opportunities in the open RTB as point processes, and show how it can be approximated in to a parametrized form of a Poisson point process. We also provide real data of the open RTB in Section 4 to illustrate how the ad slots in the market can be approximated to a Poisson process. With this framework, we widen the view of the problems in repeated auctions to be interpreted in a stochastic fashion, allowing analysis of variability of the quantities.

The second component is optimizing the bidding strategy for each of the bid opportunities. In Vickery Auctions, also known as second price auctions, it is known that the optimal bidding strategy for each participants is to bid at its truthful valuation under mild assumptions. Naturally, the first element of determination of bid price is the estimation of the objective, or the valuation. However as many of the ad markets moved on from second price auctions to first price auctions, it became crucial for participants of the auctions to estimate the strategies of the competitors, or equivalently to measure the market condition of the price of bid opportunities. There has been studies predicting the win rate, or the market price distribution of each inventory. Some approaches uses parametric distributions \cite{cui2011bid}, \cite{zhu2017gamma} or neural networks \cite{ren2019deep}, \cite{ghosh2019scalable} and takes in account how censored information is observed in the first price auction settings. With this new environment, rather than bidding on with their internal valuation, they have to discount their valuation for the bidding, also known as the "bid-shading". \cite{zhou2021efficient}, \cite{karlsson2021adaptive}, \cite{karlsson2021adaptive}, \cite{nedelec2019learning}, \cite{gao2022bidding} study the optimal strategies under the first price auction. In our new proposed framework, we allow diverse auction scenarios of how the winner is determined regarding the market condition, and the bid or action made by each players including the commonly used Vickrey auctions or first price auctions with reserve price.

Another perspective to note is that the market condition and the expected utility of a user can be dependent. In previous approaches, the estimation of the utility provided from each ad slot and the market condition is usually conducted separately. In estimation of the valuation of bid opportunities, often the expected market behaviour is not considered, and in the estimation of optimal bid shading, the randomness of the bid opportunities are not taken into account. In section 3, while providing a framework to understand and solve the problems in the RTB, we show that rather than estimating the market conditions and utilities separately, we need to take account of them simultaneously for the optimal strategy. We provide experimental results in Section 4 to support that consideration of the dependency result in benefit of performance.

\section{Approximation to Poisson Point Process}

\subsection{Formulation of RTB as Point Process}

Note that auctions of the RTB market occur repeatedly over time in a random fashion. Here we model auction opportunitys as a point process on a continuous timeline. Suppose there are $N$ users in the market. Let $\Xi_i, i= 1, 2, \ldots N$ denote the point process on $T$ that models the appearance of each users on the market. Then the total opportunities on the market could be expressed as the superpositioned process:
$$
\Xi = \sum_{i=1}^N \Xi_i
$$

A common approach to parametrize the point process is by using the Poisson point process. Assume that the opportunities from users are independent of each other, or in other words, have the memoryless property. It is well-known that such point process is actually a Poisson point process. Moreover assuming that the bid opportunities of users are independent of each other, since Poisson processes are closed under independent superposition, the process of total bid opportunities will also be a Poisson point process. 

However, the independence assumption might not be valid in some occasions. In the real world, the bid opportunities from a single users may not be independent, as there needs to be a minimum amount of time interval between the arrival of the user. Moreover there could be a positive correlation, as the user could be more likely to appear again if the user has already arrived in the near past.

In this section, we show how the point process can actually be approximated to a Poisson point process on short time intervals.

\subsection{Technical Preliminaries}

Let $\Gamma$ be a locally compact metric space. Let $\mathcal{H}$ be the space of finite point processes on $\Gamma$, so that $\xi \in \mathcal{H}$ is a non-negative integer-valued finite measure on $\Gamma$. Define the collection of measurable functions on $\mathcal{H}$ where $\mathcal{F}_{TV} := \{f:\mathcal{H} \to [-1,1]: f \text{ is measurable}\}$, $\mathcal{F}_{tv} := \{f:\mathcal{H} \to [-1,1]: f(\xi) = h(|\xi|) \text{ for some measurable } h: \mathbb{Z}^+ \to [-1,1]\}$. Now for two distributions $Q_1, Q_2$ defined on $\calH$, we can define the total variance distance and a pseudo metric as following:

\begin{align*}
d_{TV}(Q_1, Q_2) &= \sup_{g \in \mathcal{F}_{TV}} \left(\int g dQ_1 - \int g dQ_2\right)\\
d_{tv}(Q_1, Q_2) &= \sup_{g \in \mathcal{F}_{tv}} \left(\int g dQ_1 - \int g dQ_2\right)
\end{align*}

Now we introduce Palm processes. Consider two point processes $\Xi_1, \Xi_2$ on $\mathcal{H}$ where $X_1$ has a finite mean measure at $\nu$. Let $\alpha$ be a point on $\Gamma$, there exists a probability measure $Q_{\alpha}$ on $\mathcal{B}(\mathcal{H})$ such that
$$
\bbE[\Xi_1(B) 1(\Xi_2 \in M)] = \int_B Q_{\alpha}(M) \nu(d\alpha), \qquad B \in \mathcal{B}(\Gamma), M \in \mathcal{B}(\mathcal{H})
$$

We define $Q_\alpha$ as the Palm distribution of $\Xi_2$ with respect to $\Xi_1$ at $\alpha$. Now consider $Y_\alpha \sim Q_\alpha$, then we say $Y_\alpha$ is a Palm process of $\Xi_2$ with respect to $\Xi_1$ at $\alpha$. Also, when $\Xi_1 = \Xi_2 = \Xi$, we call $Y_\alpha - \delta_\alpha$ the reduced Palm process of $\Xi$ at $\alpha$. The reduced Palm process can be interpreted as the conditional distribution,  given $\Xi$ has a point at $\alpha$, the distribution of rest of the point of $\Xi$.

With these definitions, we will use the following result from \cite{chen2011poisson} to approximate bid opportunities as Poisson process.

\begin{lem}
(Theorem 2.1 from \cite{chen2011poisson}) Suppose $\{\Xi_i, i\in \mathcal{I}\}$ be a collection of point processes on $\Gamma$ with mean measures $\boldsymbol\lambda_i$. Let $\Xi = \sum_{i\in \mathcal{I}} \Xi_i$ be the superposition with finite mean measure $\boldsymbol\lambda = \sum_{i\in \mathcal{I}} \boldsymbol\lambda_i$, $\lambda = \boldsymbol\lambda(\Gamma)$. Moreover, for each $i\in \mathcal{I}$ suppose there exists a neighborhood $A_i$ such that $i \in A_i$ and $\{\Xi_j : j \in A_i^c\}$ is independent of $A_i$. Then,
\begin{align}
    d_{tv}(\mathcal{L}(\Xi), Po(\boldsymbol\lambda)) &\le \frac{1-e^{-\lambda}}{\lambda} \cdot \bbE \sum_{i\in \mathcal{I}} \int_{\Gamma} \{| |V_i| - |V_{i,\alpha}|| +| |\Xi_i| - |\Xi_{i,(\alpha)}||\} \boldsymbol\lambda_i(d\alpha) \\
    d_{TV}(\mathcal{L}(\Xi), Po(\boldsymbol\lambda)) &\le \bbE \sum_{i\in \mathcal{I}} \int_{\Gamma} \{| |V_i| - |V_{i,\alpha}|| +| |\Xi_i| - |\Xi_{i,(\alpha)}||\} \boldsymbol\lambda_i(d\alpha)
\end{align}
where $\Xi^{(i)} = \sum_{j\in A_i^c} \Xi_j$, $V_i = \sum_{j\in A_i \backslash \{i\}} \Xi_j$, $\Xi_{i,(\alpha)}$ is the reduced palm process of $\Xi_i$ at $\alpha$, and $V_{i,\alpha}$ is the Palm process of $V_i$ with respect to $\Xi_i$ at $\alpha$ where $\Xi^{(i)}+V_{i,\alpha} + \Xi_{i, (\alpha)} + \delta_\alpha$ is the Palm process of $\Xi$ with respect to $\Xi_i$ at $\alpha$. 
\end{lem}

Assuming the independence of $\Xi_i$'s, $A_i = \{i\}$ and therefore we have a simplified result:

\begin{cor}{\label{cor:tvbound}}
(Corollary 2.2 from \cite{chen2011poisson}) Suppose $\{\Xi_i, i\in \mathcal{I}\}$ be a collection of independent point processes on $\Gamma$ with mean measures $\boldsymbol\lambda_i$. Let $\Xi = \sum_{i\in \mathcal{I}} \Xi_i$ be the superposition with finite mean measure $\boldsymbol\lambda = \sum_{i\in \mathcal{I}} \boldsymbol\lambda_i$, $\lambda = \boldsymbol\lambda(\Gamma)$. Then,
\begin{align}
    d_{tv}(\mathcal{L}(\Xi), Po(\boldsymbol\lambda)) &\le \frac{1-e^{-\lambda}}{\lambda} \cdot \bbE \sum_{i\in \mathcal{I}} \int_{\Gamma} | |\Xi_i| - |\Xi_{i,(\alpha)}|| \boldsymbol\lambda_i(d\alpha) \\
    d_{TV}(\mathcal{L}(\Xi), Po(\boldsymbol\lambda)) &\le \bbE \sum_{i\in \mathcal{I}} \int_{\Gamma} | |\Xi_i| - |\Xi_{i,(\alpha)}|| \boldsymbol\lambda_i(d\alpha)
\end{align}
where $\Xi_{i,(\alpha)}$ is the reduced palm process with respect to $\alpha$. 
\end{cor}

The structure of $A_i$'s assume the local dependency. For example, we could assume there is a network-like graph structure between users to take account for the dependency between users. There could be a positive dependency, as a user induces other users to appear on the market, or using the more abstract interpretation for 'users', if there are two 'id's that is used by the same person with different identification, the two id's cannot appear at the same time, which leads to a negative dependency. Various assumption could be made on this part, but for this section, we will assume that the point process of different users are independent of each other. 

\subsection{Bound on Total Variation Distance}

Using the results, we can establish a total variance distance in our specified context.

\begin{prop} \label{prop:tvbound}
Assume that $\bbE|\Xi_i| = \lambda_i$ and $\bbE|\Xi_{i,(t)}| \le r_i$ for every $i$, $t\in T$. Let $\lambda = \sum_i \lambda_i$. For some $\delta_1, \delta_2 >0$, define
$$
\alpha = \frac{\sum_i \lambda_i 1(\lambda_i > \delta_1)}{\lambda} , \quad \beta = \frac{\sum_i \lambda_i 1(r_i > \delta_2)}{\lambda}
$$
Also assume $\lambda_i \le L$ and $r_i \le R$. Then we have the bound
$$
d_{tv}(\Xi, Po(\boldsymbol\lambda)) \le L\alpha + R\beta + \delta_1 + \delta_2
$$
\end{prop}

Proofs of the results are provided in the Appendix. Intuitively, $L$ refers to the maximum number of expected counts over a fixed time window from a single user. Note that there could be a small portion of users who appear very frequently on the market, whereas most other users rarely appear on the market. $\delta_1$ accounts for the 'common' bound for the expected counts of a user, and $\alpha$ accounts for the proportion of 'uncommon' users who appear frequently. 

$r_i$ refers to the expected counts conditioned that user $i$ has already appeared on the market at time $\alpha$. Similar to $L$, $\delta_1$ and $\alpha$, $R$ refers to the maximum, $\delta_2$ refers to the 'common' bound, $\beta$ refers to the proportion of 'uncommon' users who does not satisfy the bound.

Note that this bound will only be valid when there exists $\delta_1, \delta_2$ with low values of $\alpha, \beta$. One assumption we can make is that a user appears on the given timeline $T$ only once. Then we have $\bbE|\Xi_{i,(t)}| = 0$, which gives $R = 0, \delta_2 = 0$ and $\beta = 0$. 

Another assumption we can make is that $\lambda_i \le l |T|$ for some constant $l$. This means the the expected count from a user has some bound proportional to the size of the time interval $T$. Setting $\delta_1 = l|T|$ gives $\alpha = 0$, so under the two assumptions we have the following result.

\begin{prop}
Assume $\lambda_i \le l |T|$ for some constant $l$. Then,
$$
d_{tv} (\Xi, Po(\boldsymbol\lambda)) \le l |T|
$$
Moreover for the actual total variance distance,
$$
d_{TV} (\Xi, Po(\boldsymbol\lambda)) \le \lambda l |T|
$$
\end{prop}

Note that this bound diminishes to $0$ as we take shorter time interval $T$. 

\subsection{Formulation of RTB with Contexts}

To characterize the users we introduce the concept of \textit{contexts}. Let $(\mathcal{E}, \mathcal{F}, \mu)$ be a measure space, where $\mathcal{E}$ denotes the space of contexts, and each users have distinct context $e_i \in \mathcal{E}$. The context can refer to any characteristics of the opportunities, starting from natural characteristics such as demographic information about the user to very specific characteristics of the user, such as probability distribution of the user's behaviour after showing a certain ad.

Now consider arbitrary measurable set $B \in \mathcal{F}$. The point process of users with context in $B$ will be the superpositioned processes with context in $B$,
$$
\Xi_B := \sum_{i=1}^N 1(e_i\in B) \Xi_i
$$

where $\Xi_\mathcal{E} = \Xi$. Note that $\Xi_B(\cdot)$ is eventually $\Xi(\cdot \cap B)$. Moreover, we can extend this process to a point process on $\mathcal{E} \times T$,
$$
\Xi_B' := \sum_{i=1}^N 1(e_i \in B) \cdot \delta_{e_i} \otimes \Xi_i
$$

Where $\delta_{e_i}$ denotes the point measure on $e_i$, $\delta_{e_i}(B) = 1(e_i \in B)$. Denoting $\Xi_\mathcal{E} = \Xi'$, $\Xi_B'(\cdot) = \Xi'(\cdot \cap B)$. Note that the number of users with context $B$ will be $N_B = \sum_{i=1}^N 1(e_i\in B)$. Now we define the mean measure on $T$,
$$
\boldsymbol\lambda_B := \sum_{i=1}^N 1(e_i \in B) \boldsymbol\lambda_i
$$

where $\boldsymbol\lambda_\mathcal{E} = \boldsymbol\lambda$ and $\boldsymbol\lambda_B(\cdot) = \boldsymbol\lambda(\cdot \cap B)$. Or on $\mathcal{E}\times T$,
$$
\boldsymbol\lambda'_B := \sum_{i=1}^N 1(e_i \in B) \cdot \delta_{e_i} \otimes \boldsymbol\lambda_i
$$

where $\boldsymbol\lambda'_\mathcal{E} = \boldsymbol\lambda'$ and $\boldsymbol\lambda'_B(\cdot) = \boldsymbol\lambda'(\cdot \cap B)$.

\subsection{Moments of  Functions}

Now we assume that our quantity of interest is the sum of a function of contexts from a certain set of contexts in other words:
$$
\int h(e) d\Xi_B'(e,t)
$$

This could be any quality of interest. If $h$ is a indicator function, this quantity will be the count of opportunities with certain conditions. If $h$ is the expected conversion given the context, the aggregated quantity will be the total expected number of conversion. As the function could be arbitrary as long as it is summable, the framework gives a flexibility to handle different quantities.

First we show that if our interest is on the expectation of this quantity, it is sufficient to check the mean measure.

\begin{align*}
    \bbE \int h(e) d\Xi'_B(e,t) &= \sum_{i=1}^N 1(e_i \in B) \cdot h(e_i) \int d\Xi_i'(e,t) \\
    &= \sum_{i=1}^N 1(e_i \in B) \cdot h(e_i) \boldsymbol\lambda_i(T) 
\end{align*}

So the expectation of the quantity of interest only depends on the mean measure, so we do not have to specify the exact distribution of the point process, but only the mean measure, i.e., intensity. Now we look at the higher moments. The following result show that we can approximate higher moments by a Poisson process if the time interval is short enough.

\begin{prop}{\label{prop:secondmomentbound}}
    Suppose $|h| \le M$ for some $M>0$ and there exists $\Delta t, l$ such that $\bbE|\Xi_i| \le \l \Delta t$ for any time interval shorter than $\Delta t$, and also assume that $\Delta t$. Then,
    $$
    | \bbE f(\Xi_B)^2 - \bbE f(P_B)^2| = \tilde{\mathcal{O}}(\Delta t)
    $$
    where $\tilde{\mathcal{O}}(x)$ indicates that it is less than equal to order of $x$ ignoring logarithm terms.
\end{prop}

Similar results hold for moments higher than $2$.

\subsection{Cox Process Approximation}

Note that the Poisson approximation from the previous section requires strong assumption of the bid opportunity process, either the time interval is short enough, or the conditional expectation of opportunities given that there already has been a bid opportunity in that time window is small enough. Though these assumption may be reasonable under certain circumstances, it cannot be applied in general.

Moreover, note that the independence of counts in different time intervals cannot be guaranteed. One way to overcome this problem is to use the doubly stochastic Poisson process, or the Cox Process. The Cox process, unlike the Poisson process assumes the underlying mean measure is random, in other words, there exists a random measure $\Lambda$ such that the process is a Poisson process conditioned on $\Lambda$. This way, we can take account for the dependence of counts between time intervals, and even explain overdispersion of counts. In this section we suggest some potential models that can be used for this.

\subsubsection{Shot Noise Cox Process}

Suppose each users appear on the market, or open up the app and gives out a random number of opportunities centered at some time point. Let $\Phi_i$ be a Poisson process on $T \times (0,\infty)$ with mean measure $\zeta_i$. Let $k(c,\cdot)$ be some kernel on $T$ determined by $c\in T$. Let $\Lambda_i$ be random measure on $T$ defined as
$$
\Lambda_i = \sum_{(c,\gamma)\in \Phi_i} \gamma k(c,\cdot) 
$$

We can interpret $\Phi_i$ as the process that models the user's appearance, with center $c$ and $\gamma$ as the number of expected opportunities upon the appearance, with kernel $k$. Let $\Xi_i$ be the cox process directed by $\Lambda_i$. This is called the Shot Noise Cox Process(SNCP) \cite{moller2003shot}. Then we can define a cox process of all the users as the superposition $\Xi = \sum_{i=1}^N \Xi_i$. Note that SNCP is closed under independent superposition so $\Xi$ is another Shot Noise Cox Process directed by
$$
\Lambda = \sum_{(c,\gamma) \in \Phi} \gamma k(c,\cdot)
$$

where $\Phi$ is a Poisson process with mean measure $\zeta = \sum_i \zeta_i$, as we assume $\Phi_i$'s are independent. Note that the process can be also interpreted as a cox process on $\mathcal{E}\times T$,
$$
\Xi' = \sum_{i=1}^N \delta_{e_i} \otimes \Xi_i
$$

which is a Cox process on $\mathcal{E}\times T$ directed by
$$
\Lambda' = \sum_{i=1}^N \delta_{e_i} \otimes \Lambda_i
$$

\subsubsection{Log Gaussian Cox Process}

Although Shot Noise Cox Process is a nice model to explain the behaviour of users and the bid opportunities, It is not easy to model $\zeta_i$ or $\zeta$. Assuming that the number of user is large, a natural model we could use is the Gaussian process, taking account of the central limit theorem. But the usual Gaussian process allows negative values, which is not considerable in the case of the directing measure for Cox processes. So instead, we assume that the directing process is a exponential of the Gaussian process, in other words we assume the log Gaussian process for the directing measure. This is actually the Log Gaussian Cox Process (LGCP) \cite{moller1998log}.

The pros of using LGCP is we can parametrize the dependency structure with just the covariance, thus making parametric estimation and prediction easier. Also, it consists of only the first moment and the second moment, so we do not have to make inference about complex structure such as $\Phi$ in SNCP. Also, predictions using conditional expectation can be made with much less effort.

Now assume $\xi(t)$ is a Cox process directed by $\Lambda(t)$, where $\Lambda(t)$ is a log-Gaussian process with mean intensity $\mu(t)$ and covariance structure $C$, that is, $\log \frac{\Lambda(t)}{\mu(t)}$ is a centered Gaussian process with covariance structure $C$. Now assume the discretized time window. Then we have some known properties of the first and second moment.

\begin{prop}{\label{prop:lgcp}}
    Suppose $\xi_(t)$ is a Cox process directed by $\Lambda(t)$ where $\log\frac{\Lambda(t)}{\mu(t)}$ is a centered stationary Gaussian process on a discretized time interval with one parameter covariance function $C$. Denote $\sigma^2 = C(0)$ and $\rho(t) = C(t)/C(0)$. Then,
    \begin{enumerate}
        \item $\bbE \xi(t) = \mu(t) \cdot \exp(\frac{1}{2}\sigma^2)$
        \item $\Var(\xi(t)) = \mu(t) \cdot \exp(\frac{1}{2}\sigma^2) + (e^{\sigma^2} - 1) e^{\sigma^2} \mu(t)^2$
        \item $\Cov(\xi(t_1), \xi(t_2)) = (e^{\sigma^2 \rho(|t_1-t_2|)} -1) e^{\sigma^2} \mu(t_1) \mu(t_2)$, when $t_1 \ne t_2$
    \end{enumerate}
\end{prop}

\section{Characterization of Bid Opportunity Process}

Note that from section 2.5, the expectation of a desired quantity only depends on the intensity of the process. So on this section, we will consider the Poisson point process with the same intensity measure. So we only consider the Cox processes where $\Lambda$ is a deterministic measure.

\subsection{Formulation of RTB Continued}

For this section, let us assume that $\calE$ includes time $T$. Define a Poisson process $\eta$ with some finite intensity measure on $\mathcal{E}$, with the product measure space $(\calE, \calF_e, \mu_e)$. Note that Poisson processes with $\sigma$-finite intensity measure are uniformly $\sigma$-finite and therefore a proper point process(Cor 6.5, \cite{last2017lectures}). That is, there exists random variables $E_1, E_2, \ldots$ in $\mathcal{E}$ and $\mathbb{N}_0$-valued random variable $N$ such that almost surely
$$
\eta = \sum_{n=1}^{N} \delta_{(E_n)}.
$$
and moreover, $(E_i)$ has distribution $\Lambda/\Lambda(\mathcal{E})$, and $N$ follows a Poisson distribution with mean $\Lambda(\mathcal{E})$. We call $E_i$ the \textit{context} of bid opportunity $i$.

Let $(\calM, \calF_m)$ be some measurable space. Now we define \textit{utilities} and \textit{market conditions}. Let $K_u: \mathcal{E} \times \mathcal{B}(\bbR) \to [0,1]$ and $K_m: \mathcal{E} \times \calF_m \to [0,1]$ be transition kernels. Let $(U_1, M_1), (U_2, M_2), \ldots$ be random variables in $\bbR \times \calM$ and assume that the conditional distribution of $\{(U_i, M_i): i\le m\}$ given $N = m$ and $\{E_i : i \le m\}$ is the distribution of independent random variables with distribution $K_u(E_i, \cdot) \otimes K_m(E_i, \cdot)$, $i\le m$. We call $U_i$ the \textit{utility} of bid opportunity $i$ and $M_i \in \calM$ the \textit{market condition} of opportunity $i$. Note that by the Marking theorem(Thm 5.6, \cite{last2017lectures}), the marking process $\xi = \sum_{n=1}^N \delta_{(E_n, U_n, M_n)}$ is a Poisson process with intensity measure $\Lambda \otimes K_u \otimes K_m$. For consistency, denote $\Lambda_{E,U,M} = \Lambda \otimes K_u \otimes K_m$ and $\Lambda_{E} = \Lambda$ as the intensity measures of the variables.

The intuition behind the conditional independence of $U$ and $M$ given $E$ is that the context can be the collection of all information a bidder can get about the bid opportunity. The decisions strategy will be determined upon $E$, and therefore will be independent with $U$. Note that when an individual bidder makes bid, the bidder might not have access to every information in $E$. 

\subsection{Resolution of Contexts as sub $\sigma$-fields.}

In this subsection, we demonstrate how sub $\sigma$-fields can be used to explain the \textit{resolution} of contexts. As explained in the definition of contexts, contexts could be a very specific characteristic, and therefore might not be able to observe by the players in the auction. Instead, they observe partial context, with a certain resolution.

Suppose $\{\mathcal{F}_i : i \in \mathcal{I}\}$ is a collection of sub $\sigma$-fields of $\mathcal{F}$ on $\mathcal{E}$. Each $\mathcal{F}_i$ corresponds to a specific \textit{resolution} of the context. With resolution $\mathcal{F}_i$, for observed context $e \in \mathcal{E}$, we can only determine whether $e \in B$ for $B \in \mathcal{F}_i$. This means we only have counts of $N_B = |\Xi_B|$, $B\in \mathcal{F}_i$ and not for all $B \in \mathcal{F}$. 

Note that there exists a partial order on the collection of resolutions. If the resolution gets finer from $\mathcal{F}_1$ to $\mathcal{F}_2$, or $\mathcal{F}_1 \subset \mathcal{F}_2$, we have more information about the context $e \in \mathcal{E}$. Note that there might not be a total order. 

Another interpretation of partial context is through mapping. Suppose there exists a measurable map $\pi$ from $(\mathcal{E}, \mathcal{F})$ to $(\mathcal{E}_o, \mathcal{F}_o)$. Then $\pi^{-1}(\mathcal{F}_o)$ is a sub sigma field of $\mathcal{F}$. Intuitively, this means that if we observe a function of a context, we are observing partial information which leads to lower resolution. 

\subsection{Qualities of Interest}

 Suppose the bidder makes an action(bid) $A_i \in \calA$. Now define the \textit{win rate} function $w: \calM \times \calA \to [0,1]$. Given an action , the player wins the bid opportunity with probability $w(M_i, A_i)$, that is, $W_i \sim Bern(w(M_i, A_i))$. We call $W_i$ the \textit{win} of bid opportunity $i$.

Now consider $s: \calM \times \calA \to \bbR^+$. $s(m,a)$ is the \textit{spending} of response $r$ to a bid opportunity with market situation $m$. So the spending of bid opportunity $i$ will be $s(M_i, A_i)$. Now we define the total spending:
$$
S_{tot} = \sum_{n=1}^N W_n \cdot s(M_n, A_n)
$$
 
and define the total utility:
$$
U_{tot} = \sum_{n=1}^N W_n \cdot U_n
$$

A natural goal that comes out in this set up is to maximize profit. Let $\mu$ be the scale between utility and spending. Then we want to maximize the expected profit:
$$
\text{maximize} \ \bbE P_t := \bbE U_t - \mu \cdot \bbE S_t 
$$

Another problem we can consider is maximizing the expected total utility with respect to a budget constraint on the expected total spending:
$$
\text{maximize} \ \bbE U_t \quad \text{w.r.t.} \ \bbE S_t \le B
$$

Now we can think of the corresponding Lagrangian function:
$$
\calL := \bbE U_t - \mu \cdot \bbE S_t + \mu \cdot B
$$

So assuming $\mu > 0 $ is fixed, i.e., if the solution of the maximization problem meets the budget constraint with an equality, then the problem becomes equivalent to solving
$$
\text{maximize} \ \bbE U_t - \mu^* \cdot \bbE S_t 
$$

where $\mu^*>0$ is a constant such that this solution makes $\bbE S_t = B$. So the two optimization problem becomes almost equivalent.

\subsection{Finding the Optimal Strategy Under Limited Information}

Consider a measurable space $(\mathcal{E}_o, \mathcal{F}_o)$ with a measurable map $\pi: \mathcal{E} \to \mathcal{E}_o$. Note that $\pi_c(e,u,m) := (\pi(e),u,m)$, a map from $\mathcal{E} \times \bbR \times \calM$ to $\mathcal{E}_o \times \bbR \times \calM$ is also measurable. By the Mapping theorem(Thm 5.1, \cite{last2017lectures}) the process
$$
\pi(\xi) = \sum_{n=1}^N \delta_{(\pi(E_n), U_n, M_n)}
$$
is also Poisson process, with intensity measure $\Lambda_{E^o,U,M} :\Lambda_{E,U,M} \circ \pi_c^{-1}$. We call $E^o_i := \pi(E_i)$ the \textit{observable context} of bid opportunity $i$. 

Now assume that the player's strategy is deterministic with respect to $e_o = \pi(e)$, i.e., $A_i$ is $E_i^o$ measurable and there exists a function $a:\calE_o \to \calA$ such that $A_i = a(E_i^o)$. 

Note that a Poisson process with finite intensity measure has the distribution of a mixed binomial process(Prop 3.8). Therefore we can define a random variable $(E,U,M)$ that has the sampling distribution which is equivalent to the distribution of $(E_n, U_n, M_n)$. Note that the sampling distribution is $\Lambda_{E,U,M}/\Lambda(\calE \times \bbR \times \calM)$. Also note that $W_n, E^o_n, A_n$ is a function of $(E_n, U_n, M_n)$. Therefore we can define random variables $(W, E^o, A)$ which has the distribution of $(W_n, E^o_n, A_n)$. Now let us express the expected total spending and total cost with respect to the kernels and intensities.
\begin{align*}
    \bbE S_{tot} &= \bbE N \cdot \bbE [W \cdot s(M, A)] & (\because \text{ Wald's identity}) \\
    &= \Lambda(\mathcal{E})  \cdot \bbE[\bbE[s(M,A) \cdot W | M,A]] \\
    &= \Lambda(\mathcal{E}) \cdot \bbE[s(M,A) \cdot \bbE[W|M,A]] \\
    &= \Lambda(\mathcal{E}) \cdot \bbE[s(M,A) \cdot w(M,A)] \\
    &= \int_{\calE\times \bbR \times \calM} s(m,a) \cdot w(m,a) d\Lambda_{E,U,M}
\end{align*}

\begin{align*}
    \bbE U_{tot} &= \bbE N \cdot \bbE[W \cdot U] \\
    &= \Lambda(\mathcal{E}) \cdot \bbE[\bbE[ U \cdot W | U,M,A]] & (\because \text{ Wald's identity}) \\
    &= \Lambda(\mathcal{E}) \cdot \bbE[U \cdot w(M,A)] \\
    &= \int_{\calE\times \bbR \times \calM} u \cdot w(m,a) d\Lambda_{E,U,M}
\end{align*}

Now the Lagrangian with respect to the optimization problem will be
\begin{align*}
    \mathcal{L} &= \bbE U_{tot} + \mu(B - \bbE S_{tot}) \\
    &= \mu B +\Lambda(\calE) \bbE[U\cdot w(M,A) - \mu \cdot s(M,A)\cdot w(M,A)] \\
    &= \mu B +\Lambda(\calE) \bbE[\bbE[U\cdot w(M,a(E^o)) - \mu \cdot s(M,a(E^o)) \cdot w(M,a(E^o))|E^o]]
\end{align*}

The goal of the optimization problem is to find $a: \calE_o \to \calA$ that maximizes $\calL$, which reduces to solving
$$
\text{maximize}_{a\in \calA} \calL(a|E^o) = \bbE[U\cdot w(M, a) - \mu\cdot s(M,a)\cdot w(M,a) | E^o]
$$
$\mu_o$ almost everywhere. Then it is sufficient to solve
$$
\text{maximize}_{a\in \calA} \calL(a|E^o = e^o) = \bbE[U\cdot w(M, a) - \mu\cdot s(M,a)\cdot w(M,a) | E^o = e^o]
$$
for every $e^o\in \calE_o$. 

Considering Slater's condition, we have the following result. 
\begin{thm}[General Optimal Action for Maximizing Utility with Budget Constraint]
    Suppose opportunity triplets $(E,U,M)$ of context $E\in \calE$, utility $U \in \bbR$, market condition $M\in \calM$ occur according to a Poisson process with intensity measure $\Lambda_{E,U,M} = \Lambda \otimes K_u \otimes K_m$. Assume the player observes partial context $E^o = \pi(E)$ and makes actions $A = a(E^o)$. Suppose the player wants to solve the optimization problem
    \begin{equation} \label{eqn:profit}
    \textup{maximize}\ _{a:\calE^o \to \calA} \calL(a) := \ \bbE U_{tot} - \mu \cdot \bbE S_{tot}
    \end{equation}
    Then the optimal action is given as
    \begin{equation} \label{eqn:opt_action}
    a^*(e^o) = \argmax_{a\in \calA} \bbE[U \cdot w(M,a) - \mu\cdot s(M,a) \cdot w(M,a) | E^o = e^o]
    \end{equation}
    Now suppose the player wants to solve the constrained problem
    \begin{equation} \label{eqn:budget}
    \textup{maximize}\ _{a:\calE^o \to \calA} \ \bbE U_{tot} \quad \text{subject to} \quad \bbE S_{tot} \le B
    \end{equation}
    If there exists $\mu>0$ such that for actions as in \ref{eqn:opt_action}, the budget equality is met i.e.,
    \begin{align*}
    B &= \Lambda(\calE) \cdot \bbE[(s(M,a^*(E^o)) \cdot w(M, a^*(E^o))]\\
    &= \int_{{\calE}^o \times \bbR \times \calM} s(m, a^*(e^o)) \cdot w(m, a^*(e^o)) d\Lambda_{E^o,U,M}
    \end{align*}
    Then $a^*$ is indeed the optimal action. If there is no such $\mu$, the problem becomes equivalent to \ref{eqn:profit} with $\mu = 0$, i.e., maximizing the utility without the budget constraint.
\end{thm}

Note that this is in form of a static optimization, so the strategy(or policy) of which action to take per given observable context $e^o$ is pre-determined and does not change over time. Dynamic decision making with respect to the actual auction results can also be studied with the proposed framework. For example, in the natural example of where the player gets to make actions with respect to previous auction results and previous utilities, $A_i$ would be $(E^o_1, U_1, W_1, \ldots, E^o_{i-1}, U_{i-1}, W_{i-1}, E^o_i)$-measurable. Different scenarios of information can be considered, for example, $U_i$ might only be observable if $W_i = 1$, that is, the player can observe $U_i$ only if they won the auction, or $M_i$ might be observable if the auctioneer agrees to open the auction competitors actions to participants after the determination of the winner. While the proposed framework provides background for these scenarios, the case where actions are made via $E^o_i$ will only be considered in the rest of the work.

\subsection{Optimal Bidding in Auctions}

Now let us consider the practical auction scenarios. Note that the action player takes is in form of a bid of a positive real number i.e., $A \in \mathcal{A} = \bbR^+$. Also assume that the derivative of $w, s$ with respect to $a$ is well defined and from here and below assume that $\calL$ is concave. Taking the derivative with respect to $a$, the problem becomes equivalent to solving
$$
0 = \bbE \left[ U\cdot \frac{\partial w(M,a)}{\partial a} \Big| E^o = e^o \right] - \mu \cdot \bbE\left[ \frac{\partial (s(M,a) \cdot w(M,a))}{\partial a} \Big| E^o = e^o \right] 
$$

Note that in auctions with bidding, the player places a bid, and if the bid price is higher that other participants, the player wins the auction opportunity. Let $A \in \mathcal{A} = \bbR^+$ be the bid price, $M\in \calM = \bbR^+$ be the highest price among competitors, the market price. Then the win function $w$ will be $w(M, A) = 1(M\le A)$. Note that $\frac{\partial w}{\partial a}(m,a) = \delta(m-a)$, the dirac-delta function. Let us first consider the case of second price auction, where $s(M,A) = M$. Then the problem becomes solving

$$
\bbE\left[ U \cdot \delta(M-a) | E^o = e^o \right] = \mu \cdot \bbE\left[ M \cdot \delta(M-a) | E^o = e^o \right]
$$

Denote $F_{e^o}, f_{e^o}$ as the condition distribution and density of $M$ conditioned on $E^o = e^o$. Then the equation becomes
$$
\bbE[U | E^o = e^o, M = a] \cdot f_{e^o}(a) = \mu \cdot a \cdot f_{e^o}(a)
$$

\begin{cor}[Optimal Bidding Price for Second Price Auctions]
    In repeated second price auctions, the optimal strategy for maximizing expected utility with respect to expected budget constraint is bidding with price $a^*(e^o)$ according to observed context $e^o$, where $a^*(e^o)$ is the solution of
    $$
    a = \frac{1}{\mu} \cdot \bbE[U| E^o = e^o, M=a]
    $$
    for some $\mu>0$ satisfying
    \begin{align*}
        B &= \Lambda(\mathcal{E}) \cdot \bbE\left[ \int_0^{a^*(E^o)} m \cdot f_{E^o}(m) dm \right] \\
        &= \int_{{\calE}^o} \int_0^{a^*(e^o)} m \cdot f_{e^o}(m) dm \cdot d\Lambda_{E^o}
    \end{align*}
\end{cor}

If the distribution of the market price is independent of the utility, the optimal bid price will be
$$
a = \frac{\bbE[U|E^o = e^o]}{\mu}
$$

This is exactly the case of well known 'truthful biding' in Vickrey auctions. An interesting fact is that if we do not assume the independence of market price and utility, truthful bidding, even within the best of observable context, might not be optimal. In other words, even if we have accurately estimated the expected utility within the observable context, we have to take consideration of the market competition, even in the second price auctions.

In case of first price auction, $s(M,A) = A$. The problem becomes solving
$$
\bbE[U \cdot \delta_M(a) | E^o = e^o] = \mu \cdot \bbE[1(M \le a) + a \cdot \delta_M(a) | E^o = e^o]
$$

which is equivalent to
$$
\bbE[U | E^o = e^o, M = a] \cdot f_{e^o}(a) = \mu \cdot \left( F_{e^o}(a) +  a \cdot f_{e^o}(a)\right)
$$

\begin{cor}[Optimal Bidding Price for First Price Auctions]
    In repeated first price auctions, the optimal strategy for maximizing expected utility with respect to expected budget constraint is bidding with price $a(e^o)$ according to observed context $e^o$, where $a(e^o)$ is the solution of
    \begin{equation}
    \label{eqn:optimfpa}
    a + \frac{F_{e^o}(a)}{f_{e^o}(a)}= \frac{1}{\mu} \cdot \bbE[U| E^o = e^o, M=a]
    \end{equation}
    for some $\mu>0$ satisfying
    \begin{align*}
        B &= \Lambda(\mathcal{E}) \cdot \bbE\left[ a(E^o) \cdot F_{E^o}(m) \right] \\
        &= \int_{{\calE}^o} a(e^o) \cdot F_{e^o}(m) d\Lambda_{E^o}
    \end{align*}
\end{cor}

Note that in the first price auctions, the expected spending is monotone increasing with respect to the bid price, which diverges to $+\infty$. Therefore, as long as the existence of the solution to \ref{eqn:optimfpa} is guaranteed, there always exists $\mu > 0$ that satisfies the budget constraint with equality. Note that this might not be the case in general auctions, or even second price auctions. 

Let $f_{e^o}(u,m)$ denote the joint distribution of $U,M$ given context $e_o$. Then,
$$
\bbE[U|E^o = e^o, M = a] = \frac{\int u \cdot f_{e^o}(u,a) du}{\int f_{e^o}(u,a) du} = \frac{\int u\cdot f_{e^o}(u,a)du}{f_{e^o}(a)}
$$

Now consider the case where $U$ is binary i.e., $U\in \{0,1\}$. Denote the binary utility as \textit{conversion}. A well-studied example would be the case of click maximization, where the utility is given as a binary indicator of whether the user has clicked the ad or not. Let $f_{e^o,i}(\cdot)$ denote the conditional distribution of $M$ given $E^o = e^o$ and $U=i$. Then we have
$$
\bbE[U|E^o = e^o, M = a] = \frac{f_{e^o}(1,a)}{f_{e^o}(a)} = \bbE[U|E^o = e^o] \cdot \frac{f_{e^o,1}(a)}{f_{e^o}(a)}
$$

In cases where parametric forms of $f_{e^o}, f_{e^o,1}$ are given, we can numerically solve the optimality condition. For example, assume $M|E^o =e^o \sim Exp(\lambda_{e^o})$ and $M|E^o=e^o,U=1 \sim Exp(\lambda_{e^o,1})$. Let $u_{e^o}$ denote $\bbE[U|E^o = e^o]$. Then the optimality condition becomes
$$
g(a):= a + \frac{e^{\lambda_{e^o} \cdot a}}{\lambda_{e^o}} - \frac{1}{\lambda_{e^o}} - \frac{u_{e^o}}{\mu} \cdot e^{(\lambda_{e^o}-\lambda_{e^o,1})a} \cdot \frac{\lambda_{e^o,1}}{\lambda_{e^o}} = 0
$$

Note that $g(0) = -\frac{u_{e^o}\cdot \lambda_{e^o,1}}{\mu\cdot \lambda_{e^o}}<0$. Also the derivative is given as
$$
g'(a) = 1 + e^{\lambda_{e^o}\cdot a} - \frac{u_{e^o}}{\mu} \cdot \frac{(\lambda_{e^o}-\lambda_{e^o,1})\lambda_{e^o,1}}{\lambda_{e^o}} e^{(\lambda_{e^o}-\lambda_{e^o,1})a}
$$

If $\lambda_{e^o,1} > \lambda_{e^o}$, then $g'(a)\ge 2 >0$ for all $a>0$, and $g(a)=0$ has a unique solution. Also if $\frac{\lambda_{e^o} u_{e^o}}{2\mu} <1$, then
\begin{align*}
    g'(a) &=  1 + e^{\lambda_{e^o}\cdot a} - \frac{u_{e^o}}{\mu} \cdot \frac{(\lambda_{e^o}-\lambda_{e^o,1})\lambda_{e^o,1}}{\lambda_{e^o}} e^{(\lambda_{e^o}-\lambda_{e^o,1})a} \\
    &\ge 1 + e^{\lambda_{e^o}\cdot a}  - \frac{u_{e^o}}{\mu} \cdot \frac{\lambda_{e^o}}{2} e^{(\lambda_{e^o}-\lambda_{e^o,1})a} \\
    & \ge 1 + e^{\lambda_{e^o}\cdot a} - e^{(\lambda_{e^o} - \lambda_{e^o,1})a} \\
    & \ge 1 > 0
\end{align*}

so $g(a) =0$ has a unique solution. Restating this gives the following Lemma.
\begin{lem}
    Suppose $U$ is a binary random variable, $M$ is exponentially distributed with mean $1/\lambda_{e^o}$ and exponentially distributed with mean $1/\lambda_{e^o,1}$ conditioned on $U=1$. Suppose either $\lambda_{e^o,1}>\lambda_{e^o}$ or $\frac{\lambda_{e^o}u_{e^o}}{2\mu} <1$. Then Eqn \ref{eqn:optimfpa} has a unique solution.
\end{lem}

We give simulation results in the following section regarding this setup.

\section{Experimental Results}

\subsection{Poisson Approximation}

\begin{figure}
    \centering
    \includegraphics[width=\textwidth]{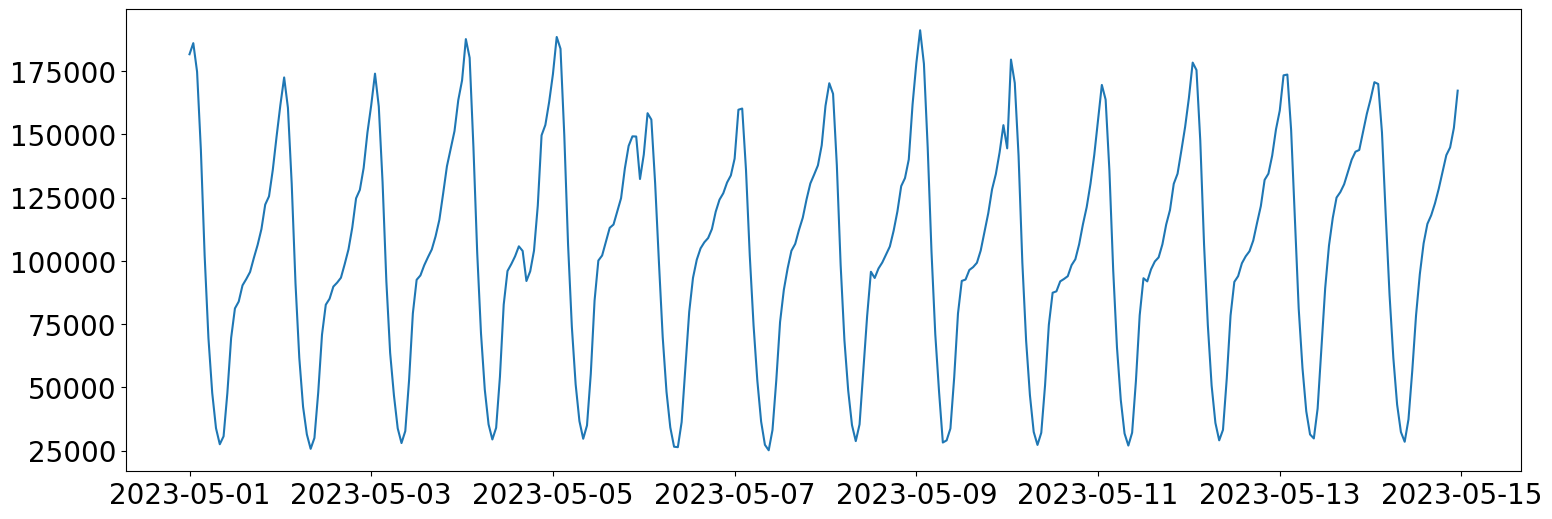}
    \caption{Number of bid opportunities per hour over two weeks from a single ad market.}
    \label{fig:daily}
\end{figure}

To validate the arguments of approximating the count process of bid opportunities to a Poisson process, we observed the number of bid opportunities of certain ad markets in the U.S. Figure \ref{fig:daily} shows the hourly number of bid opportunities from a single ad market. Note that the number of bid opportunities are highly autocorrelated, and shows a seasonal trend.

\begin{figure}
    \centering
    \includegraphics[width=\textwidth]{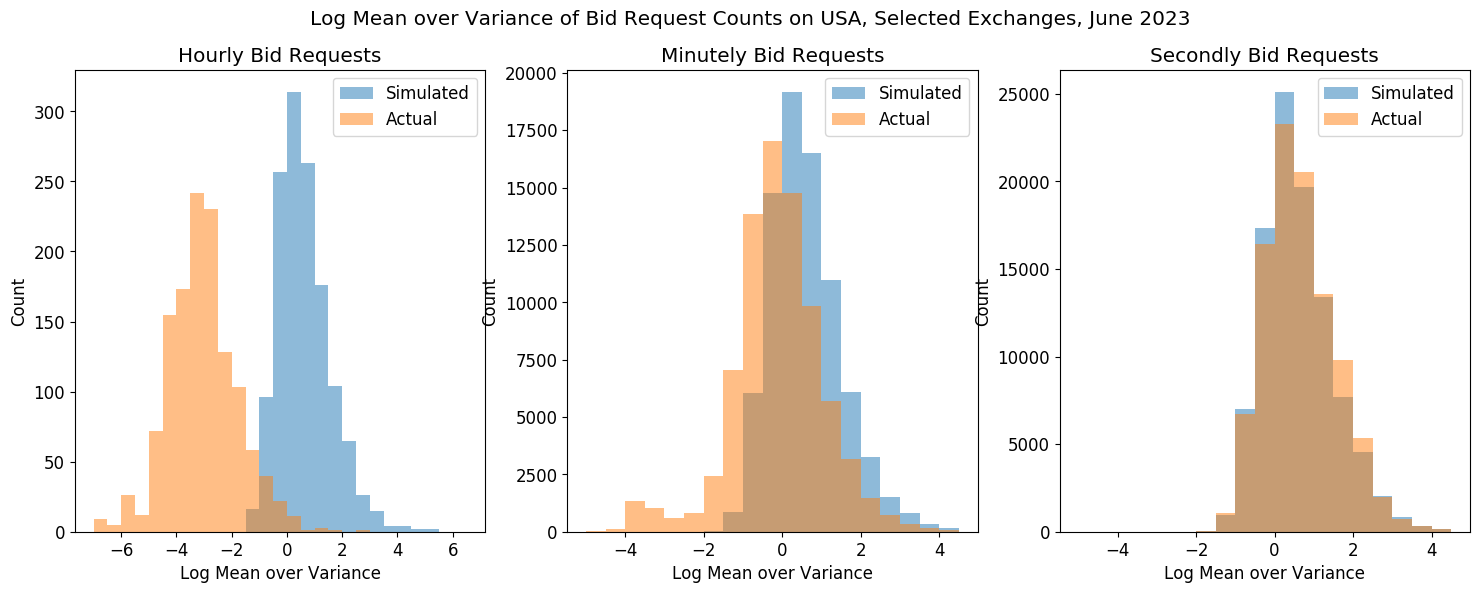}
    \caption{Histogram of logarithm of mean over variance of bid opportunities on selected 4 ad markets.}
    \label{fig:lmv}
\end{figure}

\begin{figure}
    \centering
    \includegraphics[width=0.5\textwidth]{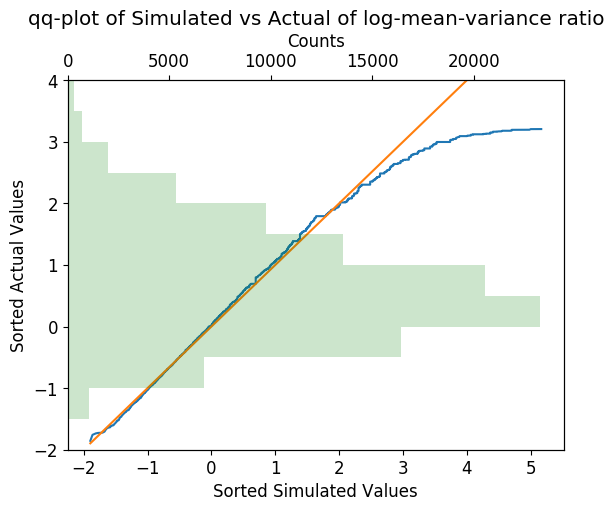}
    \caption{qq-plot and histogram of logarithm of mean over variance of bid opportunities on selected 4 ad markets, compared with the simulated distribution.}
    \label{fig:lmv_qq}
\end{figure}

Note that Poisson distributions have the same mean and variance. To illustrate that the bid opportunities in a short interval time are distributed as a Poisson distribution, we compare the distribution of the logarithm of the ratio between mean and variance on actual data and simulated data. the mean and the variance are calculated over each ad market, OS, day of week and the selected time interval over June 2023. Figure \ref{fig:lmv} shows how the distribution changes as we set the time interval from hourly to minutely to secondly. We can observe that as we have shorter time intervals, the actual distribution of the log ratio and the simulated distribution of the log ratio from the Poisson distribution coincide. Moreover, Figure \ref{fig:lmv_qq} shows the qq-plot of the simulated and the actual log mean variance ratio, and we can observe that the line is almost a straight line between $(-2,2)$ where most of the samples are located. Although the tail distribution is a bit different, it is reasonable to approximate the bid opportunity distribution as a Poisson distribution over a short time interval.

\subsection{Simulation with Dependency}

In this section, we illustrate how the optimal action suggested in Section 3 actually performs better than the strategy that doesn't consider the joint distribution. 

\subsubsection{Discrete Action and Market Condition}

As a first setup of experiment, consider the configuration where action and market condition space is discrete, and we wish to maximize profit. To demonstrate the effect of considering dependency of market condition and utility, we assume a pre-defined landscape of bid opportunities. First we assume there are $N( = 10000)$ bid opportunities with $U, M, A \in \{0, 1, \ldots, n-1\}$ with $n =20$. For each bid opportunity, we generated random joint distribution of $U$ and $M$, where
$$
\bbP(U=i, M=j) \propto \begin{cases}
Z_{ij} & \text{if } i \ne j \\
Z_{ij} +\alpha \cdot n & \text{if } i = j
\end{cases}
$$

for $i, j = 0, \ldots , n$, where $Z_{ij}$ is sampled iid from $Unif(0,1)$. This is to reflect that $U$ is likely to be equal to $M$. Note that $\alpha$ gets bigger, the dependency of utility and market price also gets larger. Then we assume that $s(m,a) = a + 1$, and we sample $w(m,a)$ from $Beta(a+1, \bbE[U|M=m]+1)$ to reflect that $a$ is the bidding price in auctions, and the spending is proportional to the bidding price, and higher the bid price is, it is more likely to win and at the same time it is less likely to win if the expected utility is large, as the competitors bids will also be high.

\begin{figure}
    \centering
    \includegraphics[width=0.5\textwidth]{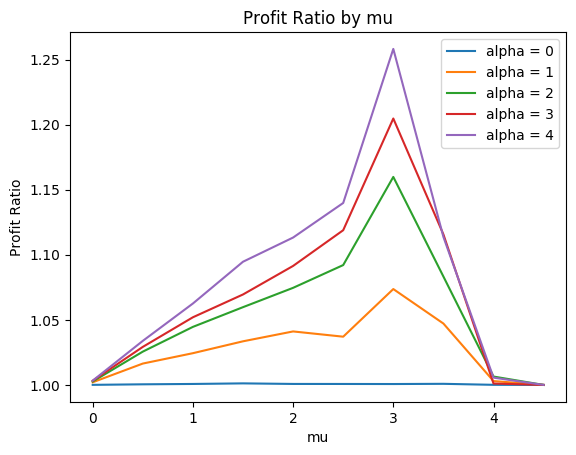}
    \caption{Change of ratio expected profit over different values of $\mu, \alpha$.}
        \label{fig:profit_by_mu}
\end{figure}

We assume that the joint distribution $\bbP_k(U = i, M = j)$ and the win probability $w_k(m,a)$ for each bid opportunities $k = 1, \ldots, N$ are known to the player. Then we compare the expected profit between when the player bids according to the optimal formula with/without considering the dependency of utility and market condition using Algorithm \ref{alg:discrete_alg}, and observe the expected profit as the scale $\mu$ changes. The results are shown on Figure \ref{fig:profit_by_mu}.

From Figure \ref{fig:profit_by_mu}, we can first see that the profit ratio is always greater than $1$, i.e., that taking account for dependency always outperforms the case where one doesn't take account for dependency. Also note that $\alpha$ takes account for the scale of dependency of the market condition and utility. We can observe as there is stronger dependency, the profit gain from considering dependency indeed increases, up to 25\%. Another fact to note is that the gap diminishes if the scale between utility and spending is extreme, when the player only needs to focus on either maximizing utility or minimizing spending.

\begin{algorithm}[H]
 \label{alg:discrete_alg}
Input: Matrix $(P)_{i,j}$ containing $\bbP(U = i, M = j)$, matrix $(S)_{m,a}$ containing spending function $s(m,a)$, matrix $(W)_{m,a}$ containing win rate $w(m,a)$, utility-spending scale $\mu$. \\

1. Calculate $\bbP(M = m) = \sum_{u} \bbP(U = u, M = m)$ and 
$$
\bbE[U|M = m] = \frac{1}{\bbP(M=m)} \sum_{u} u \cdot \bbP(U = u, M = m)
$$
for each $m\in \calM$.\\
2. Calculate the expected profit: 
$$
P(a) := \sum_m \big(\bbE[U|M = m] \cdot w(m,a) - \mu \cdot s(m,a) w(m,a)\big) \cdot \bbP(M = m)
$$
for every $a\in \calA$. \\
3. Calculate the optimal action $a^*$ by comparing $P(a)$'s:
$$
a^* = \argmax_{a\in \calA} P(a)
$$
\Return The optimal action $a^*$. 
 \caption{Choosing the Optimal Action for Discrete Space}
\end{algorithm}

\subsubsection{Exponentially Distributed Market Price}
Similar to the previous simulation, let us assume there is a pre-defined landscape of bid opportunities, but in this case in the setting of first price auction with continuous action and market price, where we wish to maximize utility with budget constraints. First we assume there are $N( = 10000)$ bid opportunities, each with a random binary utility, with its expectation distributed according to a beta distribution $p_i \sim Beta(2, 1000)$, and $U_i \sim Bernoulli(p_i)$. Next we sample $\lambda_i \sim Gamma(1, 1/p_i)$, $\log\Delta_i \sim N(\mu, \sigma^2)$ and let $\lambda_{i,1} = \lambda_i /
 \Delta_i$. We assume the marginal distribution of $M_i$ is $Exp(\lambda_i)$ and the conditional distribution conditioned on $U_i=1$ is $Exp(\lambda_{i,1})$. Note that $\Delta_i$ indicates the ratio difference between the unconditioned distribution of $M_i$ and the conditional distribution of $M_i$ conditioned on $U_i =1$. 

We assume that $p_i, \lambda_i, \Delta_i$ is known to the player and compare the sum of expected conversion when the player bids according to the optimal formula with/without considering the dependency of utility and market condition, when the budget is fixed. Searching for the value of the correct multiplier $\mu$ is performed using iterative methods, by adjusting the multiplier according to the ratio of expected spending with the given multiplier and the budget. Details are provided in Algorithm \ref{alg:spendingstabilizer}.

\begin{algorithm}[H]
 \label{alg:spendingstabilizer}
Input: Conversion probability $p_i$, distribution parameters $\lambda_{i}, \lambda_{i1}$ for each $i=1, \ldots , N$, budget $B$, initial multiplier $C_0$, allowed tolerance $\delta$.\\
0. Set $C = C_0$. \\
1. Compute $v_i = C \cdot p_i$ for each $i = 1, \ldots , N$. \\
2. Solve the equation
$$
x + \frac{e^{\lambda_i \cdot x}}{\lambda_i} - \frac{1}{\lambda_i} - C \cdot p_i \cdot e^{(\lambda_i - \lambda_{i,1})x} \cdot \frac{\lambda_{i,1}}{\lambda_i} = 0
$$
for each $i = 1, \ldots, N$, using Newton-Rhapson Method. Let $x_i$ be the solution. \\
3. Calculate the expected spending
$$
S = \sum_{i=1}^N x_i \cdot (1 - e^{-\lambda_i \cdot x_i})
$$

4. Calculate the ratio of expected spending and the budget $r = S/B$. \\
5. If $|r - 1| > \delta$, set $C \leftarrow C/\sqrt{r}$ and repeat step 1-4. If $|r-1| \le \delta$, end algorithm. \\
\Return multiplier $C$ and expected number of conversion $\sum_{i=1}^N p_i \cdot (1-e^{-\lambda_{i,1}x_i})$
 \caption{Multiplier Tuning for Fixed Budget}
\end{algorithm}

\begin{figure}
    \centering
    \includegraphics[width=0.5\textwidth]{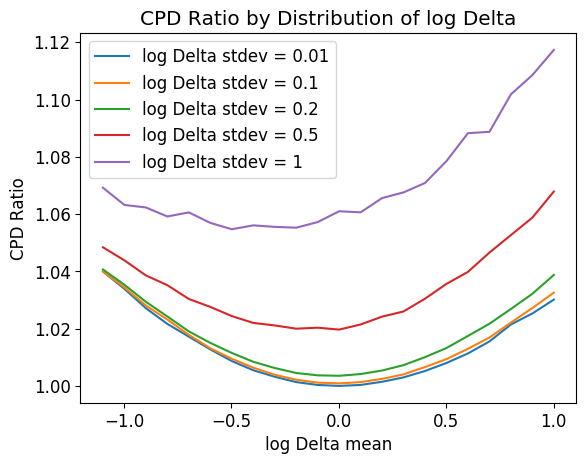}
    \caption{Change of conversion ratio on different distribution of $\log\Delta$.}
        \label{fig:cpd_by_mean}
\end{figure}

The first results on Figure \ref{fig:cpd_by_mean} are performed when the budget is fixed, but the hyperparameters that generate the difference $\log\Delta$ is changed. We compare the ratio of the expected number of conversion between the case where the formula assumes the utility and market condition is independent, and the case where the formula is based on the true dependency of utility and market condition. We can see that the formula depending on the dependency always outperforms the case where dependency is not considered, and difference of performance increases as the mean and dispersion of the dependency grows. 

\begin{figure}
    \centering
    \includegraphics[width=0.5\textwidth]{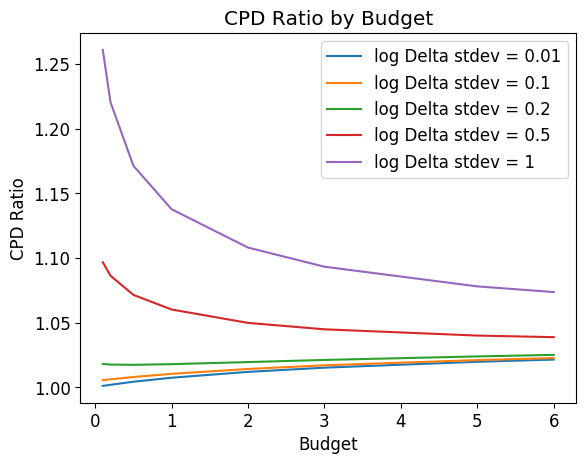}
    \caption{Change of conversion ratio on budgets and variance of $\log\Delta$.}
        \label{fig:cpd_by_budget}
\end{figure}

The second results on Figure \ref{fig:cpd_by_budget} are performed when the mean of $\log\Delta$ is fixed, but when the budget and the variance of $\log\Delta$ changes. We can observe that the new proposed method always outperforms the formula without consideration in to dependency. However we can also observe that the amount of improvement has different trends; when the variance of $\log\Delta$ is high, the difference gets larger as the budget decreases. However when the variance of $\log\Delta$ is small, i.e., the difference of market price depending on conversion is relatively stable, the amount of improvement gets larger as the budget increases.

\section{Conclusions and Future Directions}

Throughout this work, we propose a stochastic framework for modelling RTB, in the form of repeated auctions. We provide some theoretical results to provide evidence that the nature of repeated bid opportunities can be approximated to a Poisson point process. Also we provide formulas of how a player can optimize their action to maximize their objective, either to maximize profit or maximize utility with budget constraints, taking in account for the stochastic dependency structure between utility and win of each opportunity. We also provide real data and simulations to support these theories, that number of bid opportunities can be actually approximated to a Poisson distribution, and the optimal action strategy actually benefits when considering dependency.

Detailed error bounds and convergence results on other contexts are discussed in \cite{decreusefond2018stein}, \cite{decreusefond2015asymptotics} about distances between Poisson processes, Cox processes and sum of thinned point processes using Papangelou Intensities. There are also results on Gaussian Process \cite{barbour2021stein}. These methodologies using Stein's method could be used to provide more rigorous results on Cox approximation for further research on precise estimation of the process.

Another important problem DSP's has face is optimizing over multiple advertisers. While the proposed approach for optimal strategy is from each advertiser's perspective, the global optimization on aggregated utilities over multiple advertisers have not been considered. There has been studies \cite{tillberg2020optimal} on how to allocate the ad opportunities to multiple advertisers, but a more rigorous analysis on internal allocation with the proposed framework would be beneficial. 

In this paper we assumed all parameters regarding the utility and the market condition is known, but estimation of these parameters are also a major problem each player has to solve. Moreover, as the sample player observes is limited those the player wins; when the player loses the player only observes censored data. Another interesting topic in this perspective about exploration and exploitation, a classic problem in reinforcement learning. Randomized bidding with Bayesian point of view is proposed as a solution to this problem \cite{karlsson2014adaptive}, \cite{karlsson2016control}.

\newpage

\printbibliography

\newpage

\appendix

\section{Proofs}

\subsection{Proof of Proposition \ref{prop:tvbound}}
\begin{proof}
From Corollary \ref{cor:tvbound}, we have
\begin{align*}
d_{tv}(\Xi, Po(\boldsymbol\lambda)) &\le \frac{1 - e^{-\lambda}}{\lambda} \cdot \bbE \sum_i \int_{T} ||\Xi_i| - |\Xi_{i,(t)}|| \boldsymbol\lambda_i (dt) \\
&= \frac{1-e^{-\lambda}}{\lambda}  \sum_{i} (\lambda_i + r_i) \boldsymbol\lambda_i(dt) \\
&\le \frac{1}{\lambda} \sum_{i} (\lambda_i^2 + r_i \lambda_i)
\end{align*}

Also we have
\begin{align*}
    \frac{\sum_i \lambda_i^2}{\lambda} &= \frac{1}{\lambda} \left( \sum_i \lambda_i^2 1(\lambda_i > \delta_1) + \sum_i \lambda_i^2 1(\lambda_i \le \delta_1) \right) \\
    &\le \frac{1}{\lambda} \left( \sum_i L \lambda_i 1(\lambda_i > \delta_1) + \sum_i \delta_1 \lambda_i 1(\lambda_i \le \delta_1) \right) \\
    &= L\alpha + \delta_1(1-\alpha)
\end{align*}

Moreover,
\begin{align*}
    \frac{\sum_i \lambda_i r_i}{\lambda} &= \frac{1}{\lambda} \left( \sum_i \lambda_i r_i 1(r_i > \delta_1) + \sum_i \lambda_i r_i 1(r_i \le \delta_1) \right) \\
    &\le \frac{1}{\lambda} \left( \sum_i R \lambda_i 1(r_i > \delta_1) + \sum_i \delta_2 \lambda_i 1(r_i \le \delta_1) \right) \\
    &= R\beta + \delta_2(1-\beta)
\end{align*}

Joining the results, we achieve the results in the statement.

\end{proof}

\subsection{Proof of Proposition \ref{prop:secondmomentbound}}

\begin{proof}

The following lemma is a tail bound on the Poisson distribution; it can be easily derived from Chernoff's bound and the moment generating function of the Poisson distribution.
\begin{lem}
    \label{lem:PoissonTail}
    Suppose $X \sim Po(\lambda)$. Then,
    $$
    \bbP(X \ge \lambda + x) \le e^{-\frac{x^2}{\lambda+x}}
    $$
\end{lem}

Let $P_B$ a Poisson process with mean measure $\boldsymbol\lambda_B$. For brevity, denote $f(\Xi) := \int h(e) d\Xi$, $d_1 = d_{TV}(\mathcal{L}(\Xi_B), Po(\boldsymbol\lambda_B))$ and $d_2 = d_{tv}(\mathcal{L}(\Xi_B), Po(\boldsymbol\lambda_B))$. Now assume $|h| \le M$ for some $M>0$. For any $K>0$,
\begin{align*}
    |\bbE f(\Xi_B)^2 - \bbE f(P_B)^2| &\le M^2 K^2 d_1 + M^2 \cdot (\bbE[|\Xi_B|^2 \cdot 1(|\Xi_B| >K)] + \bbE[|P_B|^2 1(|P_B| > K)])
\end{align*}

Now assume there exists $\Delta t, l$ such that $\bbE|\Xi_i| \le l \Delta t$ for any time interval shorter than $\Delta t$.
\begin{align*}
    \bbE |\Xi_B|^2 &= \Var\left(\sum_i |\Xi_i| \right) + \left(\bbE \sum_i |\Xi_i| \right)^2 \\
    &= \sum_i \Var(|\Xi_i|) + \lambda_B^2 \\
    &= \sum_i \bbE|\Xi_i|^2 - \sum_i (\bbE|\Xi_i|)^2 + \lambda_B^2 \\
    &= \sum_i \bbE|\Xi_i| - \sum_i (\bbE|\Xi_i|)^2 + \lambda_B^2 \\
    &= \lambda_B^2 + \lambda_B - \sum_i (\bbE |\Xi_i|)^2
\end{align*}

where the summation over $i$ is done over $\{i: e_i \in B\}$. Since $\bbE|P_B|^2 = \lambda_B^2 + \lambda_B$, 
$$
\bbE |\Xi_B|^2 - \bbE |P_B|^2 = \sum_i (\bbE |\Xi_i|)^2 \le \lambda_B l \Delta t
$$

Therefore,

\begin{align*}
    \bbE[|\Xi_B|^2 \cdot 1(|\Xi_B| \le K)] &= \sum_{k=1}^K k^2 \bbP(|\Xi_B| = k) \\
    &= \sum_{k=1}^K k^2 \left( \bbP(k \le |\Xi_B| \le K) - \bbP(k+1 \le |\Xi_B| \le K) \right) \\
    &= \sum_{k=1}^K (2k-1) \cdot \bbP(k\le |\Xi_B| \le K) \\
    & \ge \sum_{k=1}^K (2k-1) \cdot \bbP(k \le |P_B| \le K) - K^2 d_2 \\
    &= \bbE[|P_B|^2 \cdot 1(|P_B| \le K)] - K^2 d_2
\end{align*}

which gives
$$
\bbE[|\Xi_B|^2 \cdot 1(|\Xi_B| > K)] \le \bbE[|P_B|^2 \cdot 1(|P_B| > K)] + \lambda_B l \Delta t + K^2d_2
$$

Also for $X \sim Po(\lambda)$, 
\begin{align*}
    \bbE[X^2 \cdot 1(X>K)] &= \sum_{k=K+1}^\infty (k(k-1) + k) \frac{e^{-\lambda}\lambda^k}{k!} \\
    &= \sum_{k=K+1}^\infty \left( \lambda^2 \cdot \frac{e^{-\lambda} \lambda^{k-2}}{(k-2)!} + \lambda \frac{e^{-\lambda} \lambda^{k-1}}{(k-1)!} \right) \\
    &= \lambda^2 \sum_{k=K-1}^\infty \frac{e^{-\lambda}\lambda^k}{k!} + \lambda \sum_{k=K}^\infty \frac{e^{-\lambda}\lambda^k}{k!} \\
    &= \lambda^2 \bbP(X \ge K-1) + \lambda \bbP(X \ge K) \\
    &\le (\lambda^2 + \lambda) \bbP(X \ge K-1) \\
    &\le (\lambda^2 + \lambda) e^{-K/4} & (\because \text{Lemma \ref{lem:PoissonTail}})
\end{align*}

where $K$ is large enough that $(K-\lambda-1)^2/K-1 \ge K/4$. Then we have
\begin{align*}
    |\bbE f(\Xi_B)^2 - \bbE f(P_B)^2| &\le M^2 K^2 (d_1+d_2) + M^2\lambda_B l\Delta t + 2M^2 (\lambda_B^2 + \lambda_B)e^{-K/4}
\end{align*}

Note that the minimum of the bound is achieved when $d_1 + d_2 = (\lambda_B^2 + \lambda_B) e^{-K/4}/K$. 

\begin{align*}
    |\bbE f(\Xi_B)^2 - \bbE f(P_B)^2| &\le M^2 (K^2+2K) (d_1+d_2) + M^2\lambda_B l\Delta t \\
    &\le M^2 \lambda_B l \Delta t (2K^2 + 4K + 1)
\end{align*}

Now note that
$$
(\lambda_B^2 + \lambda_B) e^{-K/4}/K = d_1 + d_2 \le 2\lambda_B l \Delta t
$$

So 

$$
\frac{e^{-K/4}}{K} \le \frac{2l\Delta t}{1+\lambda_B}
$$
which means that $K = O(\log(\frac{1}{\Delta t}))$, which completes the proof.
\end{proof}

\subsection{Proof of Proposition \ref{prop:lgcp}}

\begin{proof}
Note that $\xi(t)$ is a Poisson process conditioned on $\Lambda$ and $\Lambda(t)$ follows a log-normal distribution with parameters $(\log\mu(t), \sigma^2)$. Therefore $\bbE \Lambda(t) = \exp(\log\mu(t) + \sigma^2/2)$ and $\Var(\Lambda(t)) = (\exp(\sigma^2) - 1) \exp(2\mu(t) + \sigma^2)$. 
\begin{align*}
    \bbE \xi(t) &= \bbE[\bbE [ \xi(t) | \Lambda(t)]] \\
    &= \bbE \Lambda(t) \\
    &= \mu(t) \cdot \exp(\frac{1}{2}\sigma^2)
\end{align*}

\begin{align*}
    \Var(\xi(t)) &= \bbE[ \Var(\xi(t)| \Lambda(t))] + \Var(\bbE[\xi(t)| \Lambda(t)]) \\
    &= \bbE \Lambda(t) + \Var(\Lambda(t)) \\
    &= \mu(t) \cdot \exp(\frac{1}{2}\sigma^2) + (e^{\sigma^2} - 1) e^{\sigma^2} \mu(t)^2
\end{align*}

Also note that $\Lambda(t_1), \Lambda(t_2)$ follows a joint log-normal distribution. Therefore $\Cov(\Lambda(t_1), \Lambda(t_2)) = (\exp(\sigma^2 \rho(|t_1-t_2|)) - 1) \exp(\log\mu(t_1)+\log\mu(t_2) + \sigma^2)$. Therefore for $t_1 \ne t_2$,
\begin{align*}
    \Cov(\xi(t_1), \xi(t_2)) &= \bbE[\Cov(\xi(t_1), \xi(t_2)| \Lambda)] + \Cov(\bbE[\xi(t_1)| \Lambda] + \bbE[\xi(t_2)| \Lambda]) \\
    &= \Cov(\Lambda(t_1), \Lambda(t_2)) \\
    &= (e^{\sigma^2 \rho(|t_1-t_2|)} -1) e^{\sigma^2} \mu(t_1) \mu(t_2)
\end{align*}

\end{proof}

\end{document}